\pdfoutput=1

\documentclass[11pt]{article}

\usepackage[preprint]{acl}

\usepackage{times}
\usepackage{latexsym}
\usepackage{amsmath}

\usepackage[T1]{fontenc}

\usepackage[utf8]{inputenc}

\usepackage{microtype}

\usepackage{inconsolata}

\usepackage{graphicx}
\usepackage{subcaption}
\usepackage{amssymb}
\usepackage{algorithm}
\usepackage{algorithmic}
\usepackage{tcolorbox}
\usepackage{booktabs}
\usepackage{array}
\usepackage{makecell}
\usepackage{amssymb}
\usepackage{diagbox}
\usepackage{float}

\title{Constructing Interpretable Features from Compositional Neuron Groups
}

\author{
 \vspace{7px}
Or Shafran$^1$ ~~~~~~~ Atticus Geiger$^2$ ~~~~~~~ Mor Geva$^1$ \\ \vspace{4px}
$^1$Blavatnik School of Computer Science and AI, Tel Aviv University ~~ $^2$Pr(Ai)$^2$R Group\\
\small{\texttt{\{ordavids1@mail, morgeva@tauex\}.tau.ac.il}},
\small{\texttt{atticusg@gmail.com}}
}

\begin{document}
\maketitle

\begin{abstract}

A central goal for mechanistic interpretability has been to identify the right units of analysis in large language models (LLMs) that causally explain their outputs. While early work focused on individual neurons, evidence that neurons often encode multiple concepts has motivated a shift toward analyzing directions in activation space. A key question is how to find directions that capture interpretable features in an unsupervised manner. 
Current methods rely on dictionary learning with sparse autoencoders (SAEs), commonly trained over residual stream activations to learn directions from scratch. However, SAEs often struggle in causal evaluations and lack intrinsic interpretability, as their learning is not explicitly tied to the computations of the model. Here, we tackle these limitations by directly decomposing MLP activations with semi-nonnegative matrix factorization (\texttt{SNMF}), such that the learned features are (a) sparse linear combinations of co-activated neurons, and (b) mapped to their activating inputs, making them directly interpretable. Experiments on Llama 3.1, Gemma 2 and GPT-2 show that \texttt{SNMF} derived features outperform SAEs and a strong supervised baseline (difference-in-means) on causal steering, while aligning with human-interpretable concepts.
Further analysis reveals that specific neuron combinations are reused across semantically-related features, exposing a hierarchical structure in the MLP's activation space. Together, these results position \texttt{SNMF} as a simple and effective tool for identifying interpretable features and dissecting concept representations in LLMs.

\end{abstract}

\section{Introduction}
A central question for mechanistic interpretability is how to decompose hidden representations of large language models (LLMs) into constituent parts \cite{mueller2024questrightmediatorhistory, geiger2024causalabstractiontheoreticalfoundation, sharkey2025openproblemsmechanisticinterpretability}. 
A natural widely-studied unit is individual neural activations. However, while individual activations often demonstrate human-interpretable concepts \cite{karpathy2015visualizing, geva-etal-2021-transformer, geva-etal-2022-transformer, choi2024automatic}, it has long been established that they participate in the representation of multiple intelligible concepts \cite{Smolensky1988a, PDP1, olah2020zoom, bolukbasi2021interpretability, gurnee2023finding}.

\begin{figure}[t]
\setlength{\belowcaptionskip}{-8pt}
    \centering
    \includegraphics[width=0.9\linewidth]{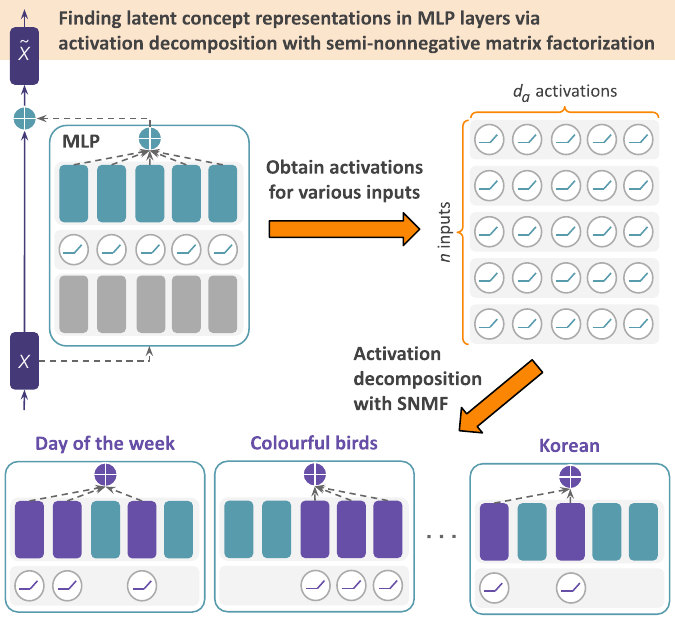}
    \caption{We find latent concept representations by decomposing MLP activations with semi-nonnegative matrix factorization (\texttt{SNMF}). Our method finds sparse compositional MLP features that align with interpretable concepts and outperform existing unsupervised methods and a strong supervised baseline at causal steering.}
    \label{fig:intro}
\end{figure}

An alternative unit of analysis that has recently gained consensus is directions in activation space \cite{elhage2021mathematical, gurnee2023finding, Nanda2023, park2024linearrepresentationhypothesisgeometry}. A key question is how to find directions encoding interpretable features in an unsupervised manner, namely, without any prior about the concepts they represent.
A large body of work has explored dictionary learning methods with sparse autoencoders (SAEs; \citealt{bricken2023monosemanticity, he2024llamascope, lieberum2024gemmascopeopensparse}) to learn directions called \textit{features}. Features are then assigned textual labels with automatic pipelines \cite{Hernandez2022, bills2023language, lee2023importanceprompttuningautomated, gur2025enhancing}.
However, while SAEs can extract interpretable features, they often fail to provide a better unit of analysis in causal evaluations (\citealt{huang2024ravelevaluatinginterpretabilitymethods,chaudhary2024evaluating, wu2025axbenchsteeringllmssimple, mueller2025mibmechanisticinterpretabilitybenchmark}; cf. \citealt{marks2025sparse}). Moreover, the learned features are not constrained to the model's representation space \cite{menon-etal-2025-analyzing, paulo2025sparseautoencoderstraineddata} nor grounded in specific model mechanisms.

In this work, we tackle the problem of finding interpretable features in LLMs by \textit{decomposing MLP activations}. Instead of learning directions from scratch, as in widely-used SAEs trained on residual stream representations, our method represents features as \textit{sparse linear combinations of existing directions in MLP weight matrices}, defined by groups of neurons (see Fig.~\ref{fig:intro}). 
To decompose MLP activations, we propose using \textit{semi-nonnegative matrix factorization} (\texttt{SNMF}; \citealt{seminmf}). We collect MLP activations with $d_a$ dimensions from a large $n$-token text, and decompose the matrix $A \in \mathbb{R}^{d_a \times n}$ of data into two matrices: $Z \in \mathbb{R}^{d_a \times k}$ of MLP features that define sparse neuron combinations, and a nonnegative coefficient matrix $Y \in \mathbb{R}_{\geq 0}^{k \times n}$. A fundamental advantage of this approach over SAEs is that the coefficient matrix $Y$ encodes which tokens contributed to the creation of a feature.  

We use texts that strongly activate features to assign textual \textit{concept labels}, and evaluate features along two axes \cite{huang-etal-2023-rigorously, wu2025axbenchsteeringllmssimple}: (1) \textit{concept detection}, measuring the ability of the feature to separate texts exemplifying the concept label and neutral texts, and (2) \textit{concept steering}, evaluating how well feature activation steers generated texts toward the concept label while preserving fluency. We find that features learned with \texttt{SNMF} outperform state-of-the-art SAEs, Gemmascope and Llamascope \cite{lieberum2024gemmascopeopensparse, he2024llamascope}, and supervised features computed with difference-in-means \cite{marks2024geometrytruthemergentlinear, rimsky-etal-2024-steering, turner2024steeringlanguagemodelsactivation} on concept steering, while performing comparably on concept detection.\looseness=-1

Further analysis shows that MLP features are compositional: recursively applying \texttt{SNMF} to the learned MLP features further composes features of fine-grained concepts (e.g., \textit{Javascript} and \textit{Python}) into features corresponding to more high-level concepts (e.g., \textit{programming}). This shows the inverse of the feature splitting phenomenon observed in SAEs, where larger SAEs have clusters of features corresponding to individual features of smaller SAEs \cite{bricken2023monosemanticity,chanin2024absorptionstudyingfeaturesplitting}. 
Moreover, semantically-related features (e.g., the different weekdays) share a core set of neurons and vary by exclusive neurons that control the promotion and suppression of fine-grained concepts.

To conclude, we propose decomposing MLP activations with \texttt{SNMF} to find interpretable features in LLMs. Our experiments show the effectiveness of this approach compared to existing unsupervised methods and a strong supervised baseline, and our analysis provides insights into how neuron compositionality constructs features in MLP layers. We publicly release the code at
\url{https://github.com/ordavid-s/snmf-mlp-decomposition}.

\section{Related Work}
\label{sec:related_work}

\paragraph{Analyzing Neurons in Transformers}
\citet{geva-etal-2021-transformer, geva-etal-2022-transformer} view the MLP output as a collection of updates to the residual stream, where updates correspond to single neurons that often encode concepts that are human-interpretable.
Alternatively, \citet{meng2022locating} considers an MLP layer as an associative memory, where all neural activations form a key that induces the layer's output. 
Recently, \citet{cao2025neurflow} used clustering to identify meaningful groups of neurons in CNNs. 
\citet{grainofsand, gurnee2023finding, OikarinenW24} used supervised methods to isolate groups of neurons in transformers that are responsible for promoting specific concepts or tasks.
Unlike prior work, we propose an unsupervised approach that identifies features via patterns in MLP activations. We view neuron combinations as causal mediators reused across inputs to represent similar concepts, extending prior work by showing that such reuse is a natural mechanism the model employs to represent concepts.

\paragraph{Concept Representations in LLMs} 
Recent work \cite{marks2024geometrytruthemergentlinear, park2024linearrepresentationhypothesisgeometry, Nanda2023} proposes the linear representation hypothesis; concept representations in language models are approximately linear, aligned with specific directions in the residual stream. However, several works challenge this view \cite{csordas-etal-2024-recurrent,park2025geometrycategoricalhierarchicalconcepts, engels2025languagemodelfeaturesonedimensionally, levy-geva-2025-language}, proposing that some concepts have meaning when viewed in multiple dimensions. These works demonstrate the presence of structured, complex concept representations.
However, they do not explain how such representations are formed or grounded in the model's underlying mechanisms. Our work applies \texttt{SNMF} to investigate how neurons form complex features, suggesting a complementary view that defines multi-dimensional concepts through their formation.

\paragraph{Finding Latent Concepts}

Supervised methods have been shown effective in finding directions in activation space that localize \cite{Geiger-etal:2023:DAS,Wu:Geiger:2023:BDAS} or steer \cite{REFT, marks2024geometrytruthemergentlinear} predefined concepts. 
However, fewer works propose unsupervised methods for finding meaningful directions. Earlier work identified salient neurons and neuron groups through probing, clustering, concept alignment, and causal analysis \citep{lakretz-etal-2019-emergence, mu_andreas_compose, dai-etal-2022-knowledge}, but these approaches generally rely on targeted analyses tied to predefined concepts or neuron-level attribution rather than recovering features directly from activation structure itself. More recently, SAEs have been used to learn directions that align with interpretable concepts \citep{bricken2023monosemanticity, cunningham2023sparseautoencodershighlyinterpretable}. Unlike SAEs, our work uses SNMF to construct features through direct matrix factorization of MLP activations, tying each feature directly to the model components responsible for forming it and representing it as a linear combination of neurons.

\paragraph{NMF and Deep Learning}
NMF has previously been used to identify interpretable concepts in deep neural networks. Prior work \cite{partsbasednmf,melaskyriazi2022deep,Fel_2023_CVPR,jourdan-etal-2023-cockatiel,oldfield2023panda} has demonstrated NMF isolates meaningful parts in images, e.g., eyes and noses, by factorizing image data. \citet{yun2023transformervisualizationdictionarylearning} applied NMF-based dictionary learning to contextualized transformer embeddings across layers, using the learned factors mainly for linguistic analysis. In contrast, we apply SNMF to MLP activations to recover features defined by groups of co-activated neurons, directly linking each feature to the model components that realize it and evaluating these features through causal steering.

\section{Decomposing MLP Activations with Semi-Nonnegative Matrix Factorization}
\label{sec:method}

We propose an unsupervised method for finding meaningful directions in LLMs, focusing on MLP layers.
Our approach is motivated by the view that the MLP output to the residual stream can be expressed as a linear combination of underlying features, each formed by a specific set of neurons. 
Their activations specify how they are linearly combined to construct concepts in a given input. Thus, inputs that express similar concepts activate similar neurons. This theory is further supported by \citet{yun2023transformervisualizationdictionarylearning} who provide evidence of the additive nature of concepts in the model's residual stream.

Following this view, we propose to identify groups of neurons representing interpretable features using semi-nonnegative matrix factorization \cite{seminmf}. 
\texttt{SNMF} factorizes a data matrix into two matrices: a matrix of \textit{features} and a nonnegative \textit{coefficient matrix} that maps between inputs and features.
Our choice of \texttt{SNMF} is motivated by \citet{partsbasednmf} who demonstrated that the nonnegativity constraints in NMF encourage parts-based, additive representations, which tend to yield interpretable features. \texttt{SNMF} retains this interpretability while relaxing the constraint for nonnegativity in the features, thus being a better fit for MLP activations, which often include both positive and negative values across architectures.

Next, we explain how we apply \texttt{SNMF} to find interpretable features in MLP layers.

\paragraph{MLP Activations}
We assume a transformer-based language model with $L$ layers, a hidden dimension $d$, and an MLP inner-dimension $d_{a}$.
An MLP layer is defined with two parameter matrices $W_K, W_V^T \in \mathbb{R}^{d_a\times d}$ and an element-wise nonlinear activation function $\sigma$:\footnote{Bias terms are omitted for brevity.}
\label{eq:mlp_out_formula}
\begin{equation}
    \texttt{MLP}(\mathbf{h}) = W_V  \sigma (W_K \mathbf{h}) := W_V \mathbf{a}
\end{equation}
where $\mathbf{h} \in \mathbb{R}^d$ is an input hidden representation and $\mathbf{a} \in \mathbb{R}^{d_a}$ is a vector of activations.\footnote{In modern LLMs, activations often go through additional gating before the output projection \cite{liu2021pay}.} Let $\mathbf{k}_i$ be the $i$-th row of $W_K$ and $\mathbf{v}_i$ the $i$-th column of $W_V$, the output of the MLP layer can be cast as a linear combination of $W_V$'s columns, each weighted by its corresponding activation:
\begin{equation}
    \texttt{MLP}(\mathbf{h}) = \sum_{i=1}^{d_a} \sigma(\mathbf{k}_i \cdot \mathbf{h}) \mathbf{v}_i = \sum_{i=1}^{d_a} a_i \mathbf{v}_i
\end{equation}

\paragraph{Decomposing Activations with SNMF}
Given neuron activations $A := [ \mathbf{a}_1 \dots \mathbf{a}_n ] \in \mathbb{R}^{d_a \times n} $ obtained from an MLP layer for a set of $n$ corresponding input representations $\mathbf{h}_1, ..., \mathbf{h}_n$, we apply \texttt{SNMF} to decompose $A$ into MLP features. Specifically, $A$ is factorized into a product of two matrices $Z \in \mathbb{R}^{d_a \times k}$ and $Y \in \mathbb{R}^{k \times n}_{\geq0}$ such that:
\begin{equation}
\label{eq:nmf}
A \approx ZY, 
\end{equation}
where $k$ is a hyperparameter defining the number of features. 

The columns of $Z$ define the representations of the \textit{MLP features}. Each MLP feature $\mathbf{z}_i \in \mathbb{R}^{d_a}$ lies in the MLP activation space and specifies a linear combination of neurons. Since the rows of $A$ are neuron activations, each MLP feature $\mathbf{z}_i$ captures a co-activation pattern across neurons, so multiplying by $W_V$ maps it to the residual stream defining a \textit{residual stream feature} $\mathbf{f}_i\in \mathbb{R}^d$: 
\begin{equation}
\mathbf{f}_i = W_V \mathbf{z}_i = \sum_{j=1}^{d_a} z_{i,j} \mathbf{v}_j 
\end{equation}
We encourage these linear combinations to be \textit{sparse}, as typically observed in MLP layers in LLMs \cite{geva-etal-2021-transformer} (see details below).
The \textit{coefficient matrix} $Y$ is restricted to nonnegative values, which specify how each input representation $\mathbf{h}_i$ is reconstructed using the MLP features. Each entry $Y_{i,j}$ indicates how strongly MLP feature $i$ contributes to sample $j$. 
Together, $Y$ and $Z$ provide a parts-based decomposition, where the original activations are reconstructed as additive combinations of the MLP features.

\paragraph{Optimization}
To learn the decomposition, we initialize the matrices $Z$ and $Y$ such that the entries of $Z$ 
are drawn from  $\mathcal{U}(0, 1)$ and the entries of $Y$ are drawn from $\mathcal{N}(0, 1)$.
We compare additional initialization strategies in \S\ref{apx:nmf_details}. Then we use the Multiplicative Updates scheme of \citet{seminmf}, which alternates between a closed-form update of $Z$ and a multiplicative update of the nonnegative matrix $Y$ to minimize the Frobenius norm between $A$ and its reconstruction $ZY$. Lastly, we impose sparsity by applying a hard winner-take-all operator to every column of $Z$, keeping only the largest $p\%$ of entries (by absolute value) and zeroing the remainder as suggested in \citet{PEHARZ201238}. We report the effect of varying $p$ in \S\ref{apx:nmf_details}.

\section{Concept Detection Evaluations}
\label{sec:interp}

We start by evaluating if the features learned by \texttt{SNMF} capture generalizable input-dependent patterns.
We show that the MLP features consistently activate in response to concept-related inputs and remain inactive for neutral inputs, suggesting that a stable group of neurons underlies each concept. Moreover, they perform comparably to those of large-scale SAEs trained on MLP outputs and better than matched-capacity SAEs, supporting their relevance to the model's input.

\subsection{Experiment}

We follow previous work \citep{paulo2024automatically, gur2025enhancing, wu2025axbenchsteeringllmssimple} and measure the ability of MLP features to discern between contexts that exemplify the concept and those that do not. Concretely, we first describe the concept captured by an MLP feature based on inputs that activate it, and then evaluate if the feature activates on inputs that were generated based on the description. 

To describe the concept captured by MLP features, we leverage GPT-4o-mini \cite{openai2024gpt4ocard} and prompt it to describe prominent patterns in the inputs that most strongly activate it \cite{bills2023language,bricken2023monosemanticity,choi2024automatic,paulo2024automatically} (see \S\ref{apx:prompt_act_neutral} for the prompt).
Then, for every description we generate five activating sentences that express the concept and five neutral sentences that do not. Next, we run each sentence through the model and record the MLP activations for every token during the forward pass.

To evaluate how well an MLP feature aligns with its description, we compute the cosine similarity between the feature and each token activation in a sentence, and take the maximum similarity per sentence.\footnote{We use cosine similarity as opposed to projection to make the scales of different methods comparable (see details in \S\ref{apx:cos_sim_vs_proj}).} We then average these maximum scores across the activating sentences to obtain $\bar{a}_{\text{activating}}$, and across the neutral sentences to obtain $\bar{a}_{\text{neutral}}$.
Lastly, we measure the log-ratio between the mean scores for the activating and neutral sentences to obtain the Concept Detection score ($S_{CD}$):
\begin{equation}
S_{CD} := \log \frac{\bar{a}_{\text{activating}}}{\bar{a}_\text{neutral}} 
\end{equation}
We use logarithm to normalize the scale for comparability across methods.
A positive log-ratio indicates greater similarity with activating sentences, where larger values indicate better separability.

We apply this evaluation to \texttt{SNMF}-learned MLP features with $k=100, 200, 300, 400$ on multiple layers of Gemma-2-2B \cite{gemmateam2024gemma2improvingopen}, Llama-3.1-8B \cite{grattafiori2024llama3herdmodels}, and GPT-2 Small \cite{radford2019language}. 
Additional details on the training of \texttt{SNMF} are in \S\ref{apx:interp_varying_k}.
As the main baseline, we use state-of-the-art SAEs, Llamascope \cite{he2024llamascope} and Gemmascope \cite{lieberum2024gemmascopeopensparse}, trained on MLP outputs (\texttt{SAE out}), i.e. $W_V \mathbf{a}$ (Eq.~\ref{eq:mlp_out_formula}). Additionally, to highlight method differences, we evaluate matched-capacity SAEs trained on MLP activations (\texttt{SAE act}) with the same dataset and number of features as \texttt{SNMF}. For additional details on training the matched-capacity baseline see \S\ref{apx:sae_training}. 
For \texttt{SNMF}, we generate concept labels from the input samples that most strongly activate it, using the coefficient matrix $Y$.
For \texttt{SAE act}, we take the feature descriptions from Neuronpedia, which are derived using autointerp \cite{paulo2024automatically} and similarly rely on the highest-activating examples. For SAEs trained on the \texttt{SNMF} dataset, we generate descriptions using their top-activating contexts with the same prompts used for 
\texttt{SNMF}.

\begin{figure}[t] 
\setlength{\belowcaptionskip}{-10pt}
\centering
\includegraphics[width=0.85\linewidth]{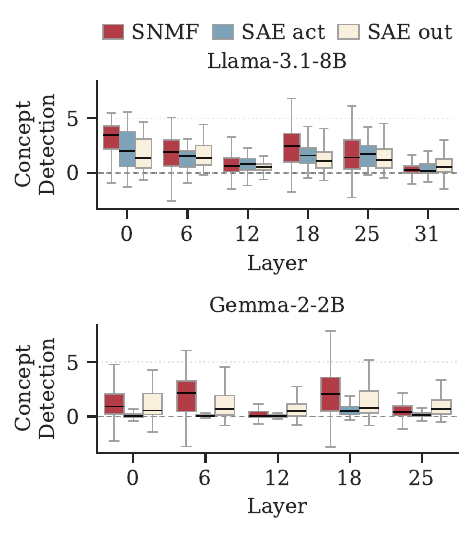}
  \caption{The distribution of concept detection scores across selected layers in Llama-3.1-8B and Gemma-2-2B for \texttt{SNMF} ($k$=100), SAE trained on MLP activations with the same dataset as \texttt{SNMF} (\texttt{SAE act}) and publicly-available SAEs trained on MLP outputs (\texttt{SAE out}).}
  \label{fig:interp_results}
\end{figure}

\subsection{Results}

Fig.~\ref{fig:interp_results} shows the Concept Detection score for \texttt{SNMF} MLP features, \texttt{SAE act} MLP features and \texttt{SAE out} features in Llama-3.1-8B and Gemma-2-2B. For \texttt{SNMF} we present the results for $k=100$, which shows similar trends to other values of $k$. Additional results showing similar trends for GPT-2 Small and comparisons across different $k$ values are in \S\ref{apx:interp_varying_k}.
Across different layers and models, most of the \texttt{SNMF} MLP features (>75\%) obtain a positive concept detection score, indicating that most \texttt{SNMF}-discovered features activate more for concept-related inputs than for neutral ones. This suggests that MLP features are meaningfully tied to the appearance of specific concepts in the input. Interestingly, we observe comparatively high scores in the first layer of both Llama-3.1-8B and GPT-2. We hypothesize that this is because the activation representations have passed through fewer attention layers so they encode fewer entangled concepts, making them easier to decompose.

When comparing methods, \texttt{SNMF} generally achieves concept detection scores that are comparable to or exceed those of existing SAEs trained on MLP outputs and a better performance compared to capacity-matched baseline SAEs trained on the \texttt{SNMF} dataset. A minority of MLP features receive negative scores, indicating they respond more strongly to neutral contexts, possibly reflecting sensitivity to noise or layer-specific biases. 

These results show that \texttt{SNMF} reliably identifies sets of neurons that correspond to human-interpretable concepts. Notably, the higher concept detection scores for \texttt{SNMF} may stem from its improved interpretability  provided by the coefficient matrix $Y$, which associates between features and their activating contexts. Table~\ref{tab:interp_examples}, presents examples of identified concepts and the tokens most strongly associated with them according to $Y$.

\begin{table}[t]
\setlength{\belowcaptionskip}{-10pt}
\setlength\tabcolsep{4pt}
\centering
\footnotesize
\resizebox{0.49\textwidth}{!}{%
\begin{tabular}{p{1.8cm}p{5.8cm}}
\toprule
\textbf{Concept} & \textbf{Top Activating Contexts} \\
\midrule
The word ``resonate'' & 
\raggedright\arraybackslash
\texttt{... nostalgia and innovation that resonates} \newline
\texttt{... captivates the eye, resonating with the} \newline
\texttt{... campaign that would resonate more strongly} \\
Implementing / establishing &
\raggedright\arraybackslash
\texttt{the police enacted curfews and employed sound cannons} \newline
\texttt{... establishing a coherent code} \newline
\texttt{schools must implement a comprehensive framework ...} \\
Historical documentation &
\raggedright\arraybackslash
\texttt{stone carvings, striving to reveal the profound stories ...} \newline
\texttt{... historical narratives preserved by the ancients} \newline
\texttt{... texts found in the dusty archives} \\
\bottomrule
\end{tabular}
}
\caption{Example concepts discovered by \texttt{SNMF} in Llama-3.1-8B, along with their top activating contexts extracted from the coefficient matrix $Y$.}
\label{tab:interp_examples}
\end{table}

\section{Concept Steering Evaluations}
\label{sec:steering}
 
In this section, we show that MLP features match and often exceed the steering performance of strong unsupervised and supervised baselines.
Together with the interpretability results, these findings show the success of our approach in isolating meaningful neuron combinations used by the model.

\subsection{Experiment}

We evaluate whether MLP features discovered by \texttt{SNMF} causally influence the model's output in a concept-consistent manner \citep{paulo2024automatically, gur2025enhancing, wu2025axbenchsteeringllmssimple}. 
To this end, we prompt the model with \texttt{``<BOS> I think that''}, while steering with a feature during inference. To control the steering strength, we perform a grid search over seven target values of KL divergence between the output logits of a standard forward pass and of an intervened pass where the feature is amplified \cite{paulo2024automatically}. As some features may suppress a concept rather than promote it, we consider both the feature $\mathbf{f}$ and its negation $-\mathbf{f}$.
For each KL divergence value and sign, we sample 8 generations, to obtain a total of 112 completions.

Next, we prompt GPT-4o-mini \cite{openai2024gpt4ocard} to score each generated continuation along two axes: \textit{concept expression} (concept score) and \textit{fluency} (fluency score).\footnote{We omit the instruct score used in \citet{wu2025axbenchsteeringllmssimple} as we evaluate base models rather than instruction tuned.} We use the evaluation prompts by \citet{wu2025axbenchsteeringllmssimple}, where each axis is scored from 0-2. For each feature, we take the best result over KL/sign combinations.
We report two scores: (a) the concept score, which evaluates how well the generated completions align with the target concept, and (b) the harmonic mean of the concept and fluency scores, which quantifies how effectively a feature can steer the model's generation toward the target concept while preserving output coherence. 
Fluency scores are less informative in isolation and are reported in \S\ref{apx:details_causality}.

We benchmark \texttt{SNMF} against Gemmascope and Llamascope (\texttt{SAE out}), the matched-capacity baseline (\texttt{SAE act}), and a strong supervised baseline, Difference-in-Means \citep[\texttt{DiffMeans};][]{wu2025axbenchsteeringllmssimple, rimsky-etal-2024-steering, marks2024geometrytruthemergentlinear, turner2024steeringlanguagemodelsactivation}. We do not include classical methods, such as PCA and ICA, as prior work \cite{wu2025axbenchsteeringllmssimple, cunningham2023sparseautoencodershighlyinterpretable} show they underperform SAEs in causal settings.
Concept scores are computed with respect to a reference feature description. \citet{gur2025enhancing} showed that input-centric descriptions are better at describing features in early layers, while output-centric descriptions are better for describing features in upper layers. Therefore, we generate descriptions for early layer features using activating inputs (as explained in \S\ref{sec:interp}). For later layers, across all methods, we use an output-centric description. For an MLP feature (\texttt{SNMF} and \texttt{SAE act}), $\mathbf{z}_i$, we calculate its projection onto the residual stream via the MLP out weights,
\begin{equation}
\mathbf{f}_i := W_V\mathbf{z}_i
\end{equation}
For features of \texttt{SAE out} and \texttt{DiffMeans} we directly use $\mathbf{f}_i$ as they were trained on MLP outputs. Next, we project the feature to the model's vocabulary space using the unembedding matrix, and generate the description based on the set of tokens corresponding to the top and bottom logits in the projection \cite{gur2025enhancing}.

\begin{figure*}[t]
\setlength{\belowcaptionskip}{-10pt}
  \centering
  \includegraphics[width=1\textwidth]{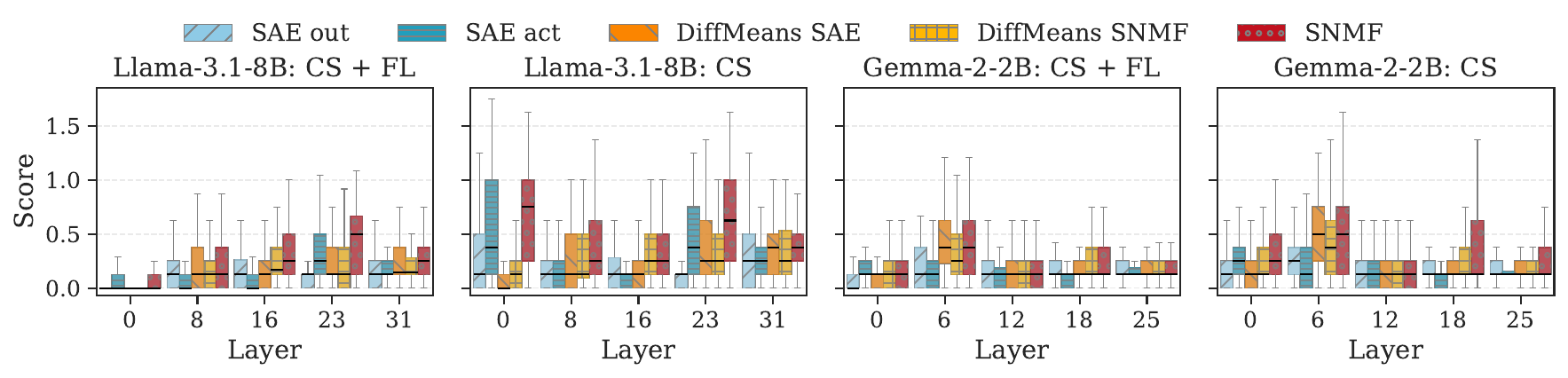}
  \caption{Causality evaluation results across layers in Llama-3.1-8B and Gemma-2-2B, using publicly-available SAEs (\texttt{SAE out}), SAE trained on MLP activations (\texttt{SAE act}), \texttt{SNMF}, and \texttt{DiffMeans} trained on concepts detected by SAE or \texttt{SNMF}. Concept Steering + Fluency (CS + FL) measure steering performance while preserving coherence, and Concept Steering (CS) quantifies how strongly the target concept is steered. Across both models \texttt{SNMF} scores often exceed both \texttt{DiffMeans} and SAEs, showing strong causal influence.}
  \label{fig:causal_results}
\end{figure*}

We apply this evaluation to \texttt{SNMF} features with $k=100$ across multiple layers in LLaMA-3.1–8B and Gemma-2–2B. For \texttt{SAE act} we use all 100 features. For \texttt{SAE out}, we randomly sample 100 features from Gemmascope and LLamascope and use the respective decoder rows. Since DiffMeans is supervised, we learn features for the concepts identified by \texttt{SNMF} and SAEs. For each concept, we use its description to generate 72 activating and 72 neutral sentences (as in \S\ref{sec:interp}) and define the feature as the difference between the average token representations for each set \citep{wu2025axbenchsteeringllmssimple}.

\subsection{Results}

Fig.~\ref{fig:causal_results} presents the results, showing varying performance across the layers of both models. Early layers tend to show lower final scores but relatively high concept scores. This discrepancy arises from the sensitivity of early-layer steering. Modifications at these layers propagate through the network, often strongly influencing model fluency and as such impacting the final score. 

Notably, \texttt{SNMF} consistently outperforms SAEs across all layers and often matches or exceeds the performance of \texttt{DiffMeans}, a supervised baseline. 
Since \texttt{DiffMeans} computes feature directions by subtracting positive and negative representations, it is susceptible to noise by unrelated concepts present in the positive samples.
This issue is especially prominent in earlier layers, where limited contextualization leaves token representations within the positive sample distinct, making their average less meaningful. In contrast, \texttt{SNMF} produces a parts-based decomposition that captures structure consistently associated with the target concept, leading to improved performance across all layers.

These results demonstrate that \texttt{SNMF} is able to isolate groups of neurons encoding concepts, supporting its viability as an interpretable unsupervised method for discovering meaningful directions through the MLP. Furthermore, the ability of MLP features produced by \texttt{SNMF} to reliably steer model outputs implies that the MLP operates using additive updates composed of interpretable neuron sets, which extends the findings of \citet{geva-etal-2022-transformer} that single MLP vectors promote concepts in the residual stream. Each MLP feature corresponds to a distinct combination of neurons whose activations affect the residual stream in ways aligned with human-interpretable concepts. 
Considering that neuron combinations activate to represent a concept, it follows naturally that a single neuron may participate in multiple concepts, and that a single concept may activate multiple neurons.

\section{Analyzing Neuron Compositionality}

Our experiments show that \texttt{SNMF} finds groups of neurons that represent interpretable features. 
A natural question that arises is \textit{how are different neurons combined to form features?}
Here, we investigate this question using the properties of \texttt{SNMF}.

\paragraph{Feature Merging via Recursive SNMF}
\label{subsec:concept_comp}

We apply \texttt{SNMF} recursively with progressively smaller values of $k$, which encourage more generic patterns \cite{hier_nmf}. Namely, we start by decomposing the activations and then recursively decompose the generated MLP features.
This process reveals subsets of neurons that are shared across features, forming a hierarchy that shows how the model additively combines groups of neurons to represent concepts.
We run this experiment on a dataset of sentences related to the concept of time units (e.g., months, weekdays, minutes) and use the coefficient matrices $Y$ from each step to identify relationships between MLP features. More details on the recursive application of \texttt{SNMF} are in \S\ref{apx:neuron_comp}.

\begin{figure}[t]
\setlength{\belowcaptionskip}{-5pt}
  \centering
  \includegraphics[width=0.98\columnwidth]{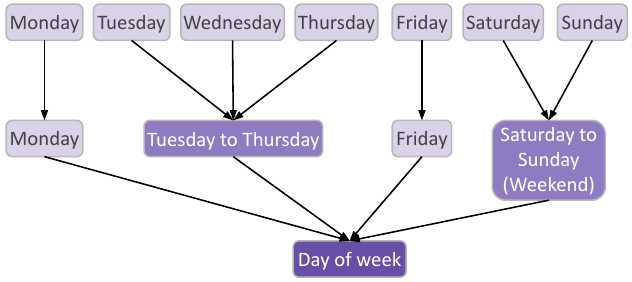}
  \caption{Example feature merging in GPT-2 Small via recursive \texttt{SNMF}, showing the concepts corresponding to the features identified at each step. 
  }
  \label{fig:weekday_decomp}
\end{figure}

Fig.~\ref{fig:weekday_decomp} illustrates the resulting hierarchy of concepts corresponding to the MLP features learned at each step. Concept labels were derived from the top activating contexts of the MLP feature according to the coefficient matrix.
For example, the \textit{weekend} node was activated for contexts mentioning Saturday, Sunday and the word ``weekend''. Overall, we see that the resulting structure progresses from groups of neurons that represent fine-grained concepts (i.e., specific weekdays) to more high-level concepts (middle/end of the week and ``day of week''). 
This feature-merging behavior is the inverse of what has been described as feature splitting in SAEs \cite{chanin2024absorptionstudyingfeaturesplitting, bricken2023monosemanticity}, where a single feature in a smaller SAE corresponds to multiple features in a larger SAE.
Additional recursive decomposition examples learned from a more general dataset are in \S\ref{apx:neuron_comp}.

\paragraph{Semantically-Related MLP Features Share Common Neuron Structures}
\label{subsec:causal_combinations}

The previous analysis suggests that certain groups of neurons contribute to semantically-related MLP features. Here, we look into these shared structures using the matrix $Z$. 
We binarize $Z$ with a per-feature threshold to produce $\bar{Z}$, where the top activating neurons of each MLP feature are set to 1 and the rest to 0.
Then, we calculate
$M := \bar{Z}\bar{Z}^T$,
where each entry $M_{i,j}$ is the number of overlapping, top activating neurons between the MLP features $i$ and $j$.

\begin{figure}[t]
\setlength{\belowcaptionskip}{-10pt}
  \centering
\includegraphics[width=0.6\columnwidth]{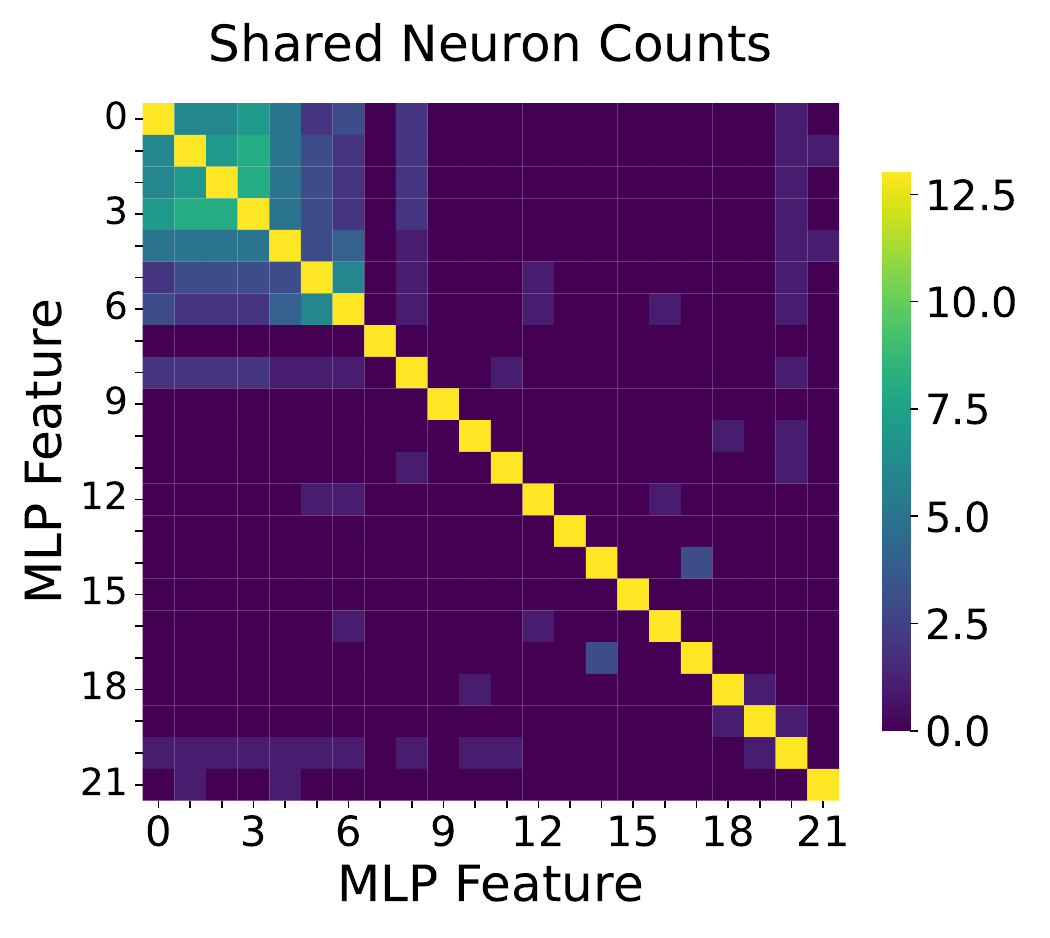}
  \caption{Shared neuron counts between MLP feature pairs. Features 0–6 correspond to weekdays (Monday-Sunday), the rest are randomly sampled from $Z$. All days share core neurons, suggesting a general day representation, while weekdays and weekends share distinct subsets, forming specialized neuron cores.}
  \label{fig:weekday_heatmap}
\end{figure}

Fig.~\ref{fig:weekday_heatmap} visualizes $M$, where the first seven rows correspond to MLP features associated with Monday through Sunday and the rest MLP features are randomly sampled from $Z$. The heatmap reveals that the model represents the concept of weekdays using a common set of neurons, as indicated by the higher shared neuron counts among the first seven rows and columns. Given the increased diagonal values, we understand that each day also activates its own additional distinct subset of neurons. Furthermore, the MLP features representing the weekend and those representing weekdays each share more neurons within their respective groups. We can see this in rows 5 and 6, which have higher shared neuron counts with each other than in rows 0 to 4 (and vice versa). Meanwhile, there is little overlap with the remaining MLP features. These patterns show that the model utilizes sets of neurons as building blocks to represent complex and hierarchical concepts. Specifically, a core set of neurons encodes the concept of a day, while additional neurons refine the representation into weekend or weekday, and further neurons produce representations for individual days.

We test this via causal interventions. Let the \textit{neuron base} be the neurons shared across all weekday MLP features, and the \textit{exclusive neurons} the unique neurons to each MLP feature. We feed the prompt \texttt{``I think that''} to GPT-2 Large, while amplifying the neuron base and, separately, the exclusive neurons of each weekday. We measure the effect of each intervention on the model's output distribution by calculating the change in logits for tokens associated with each day of the week.

\begin{table}[t]
\setlength{\belowcaptionskip}{-8pt}
\setlength\tabcolsep{3.5pt}
\centering
\footnotesize
\begin{tabular}{lrrrrrrr}
Neuron group & \rotatebox{90}{Monday} & \rotatebox{90}{Tuesday} & \rotatebox{90}{Wednesday} & \rotatebox{90}{Thursday} & \rotatebox{90}{Friday} & \rotatebox{90}{Saturday} & \rotatebox{90}{Sunday} \\
\midrule
Monday    &  \textbf{2.0} & \textcolor{red}{-0.8} & \textcolor{red}{-1.0} & \textcolor{red}{-1.2} & \textcolor{red}{-0.8} & \textcolor{red}{-1.1} & \textcolor{red}{-0.1} \\
Tuesday   & \textcolor{red}{-3.3} & \textbf{0.5} & \textcolor{red}{-2.2} & \textcolor{red}{-2.4} & \textcolor{red}{-2.4} & \textcolor{red}{-3.2} & \textcolor{red}{-2.8} \\
Wednesday & \textcolor{red}{-2.7} & \textcolor{red}{-3.2} & \textbf{2.5} & \textcolor{red}{-2.7} & \textcolor{red}{-4.3} & \textcolor{red}{-3.2} & \textcolor{red}{-2.7} \\
Thursday  & \textcolor{red}{-0.9} & \textcolor{red}{-1.0} & \textbf{0.6} & \textbf{4.5} & \textcolor{red}{-0.1} & \textcolor{red}{-1.7} & \textcolor{red}{-0.5} \\
Friday    & \textcolor{red}{-2.9} & \textcolor{red}{-2.8} & \textcolor{red}{-2.7} & \textcolor{red}{-1.6} & \textbf{1.3} & \textcolor{red}{-1.9} & \textcolor{red}{-2.6} \\
Saturday  & \textcolor{red}{-0.2} & \textcolor{red}{-0.7} & \textcolor{red}{-0.1} & \textcolor{red}{-0.7} & \textcolor{red}{-0.5} & \textbf{2.4} & \textbf{0.7} \\
Sunday    & \textcolor{red}{-0.4} & \textcolor{red}{-1.7} & \textcolor{red}{-1.6} & \textcolor{red}{-2.0} & \textcolor{red}{-0.8} & \textcolor{red}{-0.8} & \textbf{2.6} \\
\midrule
Core weekday  & \textbf{5.8} & \textbf{5.7} & \textbf{5.4} & \textbf{5.8} & \textbf{6.0} & \textbf{5.4} & \textbf{4.7} \\
\end{tabular}
\caption{Change in logits for weekday tokens (columns) when amplifying exclusive neurons tied to different weekday concepts (rows). Positive values indicate promotion; negative indicate suppression. Exclusive neurons promote their own token (diagonal) and suppress others, while core neurons (bottom row) raise all logits, encoding a general “weekday” concept.}
\label{tab:causal_effect_exclusive_neurons}
\end{table}

Table~\ref{tab:causal_effect_exclusive_neurons} shows that amplifying the neuron base promotes all weekday tokens, suggesting it encodes a high-level weekday concept. Conversely, amplifying a day's exclusive neurons promotes its associated token while suppressing the others. For example, activating Sunday's exclusive neurons suppresses the tokens for ``Monday'' through ``Saturday''. These findings demonstrate that the model builds increasingly specific concepts by using sets of neurons as compositional building blocks.

In \S\ref{apx:hierarchy}, we quantitatively evaluate hierarchies in MLP layers, showing that broader parent features often cover the causal effects of their children.

\section{Conclusion}

We introduce an unsupervised method to decompose MLP activations into MLP features, which represent how groups of neurons contribute to encoding concepts. The MLP features often align with human-interpretable concepts, performing on par with SAEs on interpretability benchmarks, while achieving better steering performance than both SAEs and a strong supervised baseline, \texttt{DiffMeans}. We further demonstrate that the model composes groups of neurons additively to represent a wide range of concepts, revealing a hierarchical organization in the hidden activation space. This behavior helps explain why the same sets of neurons contribute to multiple features, extending the functional role of neurons in the MLP. It also supports the view that the feature splitting observed in SAEs reflects the model’s use of neuron composition to construct more specific features, rather than necessarily being a byproduct of SAE training. Overall, our findings provide a new lens on how MLPs form features in LLMs and we introduce an interpretable unsupervised approach for identifying and interpreting MLP features in transformers.

\section*{Limitations}
While our work introduces a promising method for analyzing MLP activations, our benchmarks were limited to a feature count of $k<500$. Although previous research \cite{scalesnmf} has explored ways to scale NMF in different settings, we focused this paper on establishing the viability of SNMF for activation decomposition. Scaling the method to larger $k$ values requires additional engineering efforts, which we left out of scope to focus on introducing the tool and presenting new findings on concept hierarchies. Exploring the finer-grained features that may emerge at higher $k$ and testing whether the method scales effectively to thousands of features are natural directions for future work.

For evaluation, we rely on LLMs as judges. To increase confidence that using an LLM judge is appropriate in our setting, we follow standard LLM as a judge practices from prior work and additionally verify agreement with humans on our task. Specifically, we ran a human evaluation on a balanced, randomly sampled set of 50 judged examples and measured alignment with the LLM judge, finding strong agreement (Spearman $\rho = 0.8$, $p < 10^{-4}$). While this check supports that the judge behaves consistently with human judgments in our context, results may still exhibit some sensitivity to prompt design and other evaluation details.

Finally, following the original SNMF and NMF papers we use Multiplicative Updates (MU) for the optimization. However, MU can be limiting when extending the method. For future work it may be beneficial to utilize projected gradient descent, which may offer a stronger foundation by supporting regularization and potentially improving performance.

\section*{Acknowledgments}
We thank Aaron Mueller for his valuable feedback and guidance on training sparse autoencoders. This work was supported in part by the Gemma 2 Academic Research Program at Google, a grant from Open Philanthropy, the Alon Scholarship, and the Israel Science Foundation grant 1083/24.

\bibliography{custom}

\appendix
\section{Activation Decomposition with SNMF}
\label{apx:nmf_details}

In this section we describe in detail the process used to decompose MLP activations with \texttt{SNMF}. 

\paragraph{Initialization}
We compare three initialization strategies for SNMF: \textbf{Random} (our baseline), \textbf{SVD-based} \cite{BOUTSIDIS20081350}, and \textbf{K-Means} \cite{seminmf}. For Random initialization, the non-negative activation matrix \(Y \in \mathbb{R}_{\ge 0}^{k \times n}\) is sampled element-wise from \(\mathrm{Uniform}[0, 1]\), and the feature matrix \(Z \in \mathbb{R}^{d \times k}\) is drawn from \(\mathcal{N}(0, 1)\). For SVD and K-Means, we follow the procedures described in their respective references. All variants are trained using multiplicative updates.

\vspace{0.5em}
\noindent\textbf{Concept Detection.} The performance across layers for each initialization is shown below:

\begin{table}[h]
\centering
\label{tab:init-concept-detection}
\resizebox{\linewidth}{!}{
\begin{tabular}{lcccccc}
\toprule
\textbf{Init Strategy} & \textbf{Layer 0} & \textbf{Layer 6} & \textbf{Layer 12} & \textbf{Layer 18} & \textbf{Layer 25} & \textbf{Layer 31} \\
\midrule
SVD       & $2.76 \pm 1.79$ & $1.63 \pm 1.26$ & $0.94 \pm 0.92$ & $2.26 \pm 1.68$ & $2.07 \pm 1.72$ & $0.47 \pm 0.88$ \\
K-Means   & $2.55 \pm 1.51$ & $1.69 \pm 1.32$ & $0.79 \pm 0.81$ & $2.45 \pm 1.44$ & $1.58 \pm 1.58$ & $0.85 \pm 0.97$ \\
Random    & $2.99 \pm 1.55$ & $1.67 \pm 1.39$ & $0.81 \pm 0.94$ & $2.35 \pm 1.68$ & $1.89 \pm 1.72$ & $0.48 \pm 0.62$ \\
\bottomrule
\end{tabular}
}
\caption{Concept detection performance (mean \(\pm\) std) across initialization strategies.}
\end{table}

\vspace{0.5em}
\noindent\textbf{Concept Steering and Fluency.} We evaluate concept steering quality jointly with generation fluency. Results are consistent across initializations:

\begin{table}[h]
\centering

\label{tab:init-steering-fluency}
\resizebox{\linewidth}{!}{
\begin{tabular}{lccccc}
\toprule
\textbf{Init Strategy} & \textbf{Layer 0} & \textbf{Layer 8} & \textbf{Layer 16} & \textbf{Layer 23} & \textbf{Layer 31} \\
\midrule
SVD       & $0.10 \pm 0.17$ & $0.28 \pm 0.32$ & $0.32 \pm 0.29$ & $0.41 \pm 0.31$ & $0.25 \pm 0.17$ \\
K-Means   & $0.10 \pm 0.14$ & $0.29 \pm 0.33$ & $0.28 \pm 0.28$ & $0.47 \pm 0.33$ & $0.30 \pm 0.18$ \\
Random    & $0.10 \pm 0.12$ & $0.27 \pm 0.32$ & $0.31 \pm 0.29$ & $0.45 \pm 0.32$ & $0.31 \pm 0.18$ \\
\bottomrule
\end{tabular}
}
\caption{Concept steering + fluency scores (mean \(\pm\) std) across initialization strategies.}
\end{table}

\vspace{0.5em}
\noindent\textbf{Convergence Speed.} While all three methods yield similar performance, K-Means and SVD lead to faster convergence on average:

\begin{table}[h]
\centering
\label{tab:init-convergence}
\resizebox{0.75\linewidth}{!}{
\begin{tabular}{lc}
\toprule
\textbf{Init Strategy} & \textbf{Iterations to Convergence} \\
\midrule
K-Means   & $1484 \pm 1202$ \\
SVD       & $2474 \pm 2068$ \\
Random    & $3325 \pm 2975$ \\
\bottomrule
\end{tabular}
}
\caption{Average number of iterations to convergence (mean\(\pm\)std) across all Llama 3.1‑8B layers.}
\end{table}

\vspace{0.5em}
\noindent Overall, SNMF demonstrates robustness to initialization choice, both in performance and interpretability. The modest convergence benefit of SVD and K-Means makes them practical alternatives, especially in large-scale settings.

\vspace{0.5em}
\noindent\textbf{Optimization.} We also experimented with projected gradient descent during early development and observed performance comparable to multiplicative updates. Given its simplicity and widespread use in SNMF literature, we adopt multiplicative updates in our main experiments. A full optimizer comparison is left for future work.

\paragraph{Feature update}
With \(Y\) fixed we obtain \(Z\):
\[
  Z \;\leftarrow\;
  A\,Y^{\top}\!
  \bigl(YY^{\top} + \lambda I\bigr)^{-1},
\]
where \(\lambda>0\) is a small regularisation constant that prevents
ill-conditioning.

\paragraph{Activation update}
Keeping \(Z\) fixed, the non-negative activation matrix \(Y\) is updated
using the multiplicative rule derived from the \texttt{SNMF} objective:
\[
  Y \;\leftarrow\;
  Y \;\odot\;
  \sqrt{
    \frac{
      \bigl[Z^{\top}A\bigr]_{+}
      \;+\;
      \bigl[Z^{\top}Z\bigr]_{-}\,Y
    }{
      \bigl[Z^{\top}A\bigr]_{-}
      \;+\;
      \bigl[Z^{\top}Z\bigr]_{+}\,Y
    }
  },
\]
where \([X]_{+}\) (\([X]_{-}\)) denotes the element-wise positive
(negative) part of \(X\), and \(\odot\) is the Hadamard product.  
This update preserves the non-negativity of \(Y\).

\paragraph{Sparsity and Normalization}
We enforce sparsity in the feature matrix \(Z\) using a hard winner-take-all (WTA) operator: at each training step, only the top \(p\%\) of entries (by absolute value) are retained per column, with all others zeroed. Unless otherwise stated, we use \(p = 1\%\). To stabilize training, each column of the activation matrix \(Y\) is normalized to unit \(\ell_2\) norm, and the corresponding column of \(Z\) is rescaled accordingly to keep the product \(ZY\) unchanged. This method is adapted from \citet{PEHARZ201238}, who use a similar approach for sparsity in Non-negative Matrix Factorization.

\vspace{0.5em}
\noindent\textbf{Sensitivity to Sparsity.} We evaluate the effect of varying the WTA threshold (\(p \in \{1\%, 5\%, 10\%\}\)) on Llama-3.1 8B. As shown below, the \(p = 1\%\) setting used in the paper remains optimal across both concept detection and causal steering metrics:

\begin{table}[htbp]
\centering
\label{tab:wta-concept-detection}
\resizebox{\linewidth}{!}{
\begin{tabular}{lcccccc}
\toprule
\textbf{WTA \%} & \textbf{Layer 0} & \textbf{Layer 6} & \textbf{Layer 12} & \textbf{Layer 18} & \textbf{Layer 25} & \textbf{Layer 31} \\
\midrule
1   & $2.99 \pm 1.55$ & $1.67 \pm 1.39$ & $0.81 \pm 0.94$ & $2.35 \pm 1.68$ & $1.89 \pm 1.72$ & $0.48 \pm 0.62$ \\
5   & $2.60 \pm 1.40$ & $1.64 \pm 1.19$ & $0.89 \pm 0.92$ & $2.29 \pm 1.65$ & $1.75 \pm 1.46$ & $0.40 \pm 0.57$ \\
10  & $2.51 \pm 1.56$ & $1.78 \pm 1.22$ & $1.03 \pm 0.97$ & $1.86 \pm 1.62$ & $1.98 \pm 1.70$ & $0.51 \pm 0.70$ \\
\bottomrule
\end{tabular}
}
\caption{Concept detection (mean \(\pm\) std) across different WTA sparsity levels.}
\end{table}

\begin{table}[htbp]
\centering

\label{tab:wta-steering-fluency}
\resizebox{\linewidth}{!}{
\begin{tabular}{lccccc}
\toprule
\textbf{WTA \%} & \textbf{Layer 0} & \textbf{Layer 8} & \textbf{Layer 16} & \textbf{Layer 23} & \textbf{Layer 31} \\
\midrule
1   & $0.10 \pm 0.12$ & $0.27 \pm 0.32$ & $0.31 \pm 0.29$ & $0.45 \pm 0.32$ & $0.31 \pm 0.18$ \\
5   & $0.07 \pm 0.10$ & $0.23 \pm 0.27$ & $0.27 \pm 0.31$ & $0.41 \pm 0.30$ & $0.29 \pm 0.18$ \\
10  & $0.05 \pm 0.08$ & $0.23 \pm 0.27$ & $0.23 \pm 0.26$ & $0.34 \pm 0.30$ & $0.27 \pm 0.18$ \\
\bottomrule
\end{tabular}
}
\caption{Concept steering + fluency (mean \(\pm\) std) across different WTA sparsity levels.}
\end{table}

\noindent The 1\% setting provides the best trade-off between sparsity and performance, validating its selection as default in our main experiments.

\section{Cosine Similarity versus Projection}
\label{apx:cos_sim_vs_proj}
If we consider the formula for projection we get,
\begin{align}
m_{\rm proj}(S)
&= \frac{1}{|S|}\sum_{\mathbf{x}\in S} \mathbf{f}^\top \mathbf{x} \notag \\
&= \|\mathbf{f}\| \cdot \frac{1}{|S|} \sum_{\mathbf{x}\in S} \|\mathbf{x}\| \cos\theta_x
\end{align}

so for activating \(S_{\rm act}\) and neutral \(S_{\rm neu}\),
\begin{equation}
\frac{m_{\rm proj}(S_{\rm act})}{m_{\rm proj}(S_{\rm neu})}
=\frac{\sum_{x\in S_{\rm act}}\|\mathbf{x}\|\cos\theta_x}
     {\sum_{y\in S_{\rm neu}}\|\mathbf{y}\|\cos\phi_y}
\end{equation}
depends on the norms \(\|\mathbf{x}\|\), \(\|\mathbf{y}\|\).  
By contrast,
\begin{equation}
m_{\cos}(S)
=\frac{1}{|S|}\sum_{\mathbf{x}\in S}\frac{\mathbf{f}^\top \mathbf{x}}{\|\mathbf{f}\|\|\mathbf{x}\|}
=\frac{1}{|S|}\sum_{\mathbf{x}\in S}\cos\theta_{\mathbf{x}}
\end{equation}
cancels \(\|\mathbf{f}\|\) and \(\|\mathbf{x}\|\). Since both unsupervised methods compare against different activating and neutral sets, using cosine similarity makes the activating/neutral ratio directly comparable across methods.

\section{Matched-Capacity Baseline SAE Training on MLP Activations}
\label{apx:sae_training}
For a matched-capacity baseline in the causal evaluation we trained sparse autoencoders (SAEs) on MLP activations from LLaMA-3.1–8B at layers 0, 8, 16, 23, and 31 and Gemma-2–2B at layers 0, 6, 12, 18, and 25. For concept detection we additionally trained on layers 6, 12, 18, and 25 for LLaMA-3.1–8B. For training, we followed best practices as outlined in \citet{gao2024scalingevaluatingsparseautoencoders}. We trained both TopK and L1 regularized architectures across a grid of hyperparameters: three learning rates $(1\mathrm{e}{-4},\ 5\mathrm{e}{-4},\ 1\mathrm{e}{-5})$ and three sparsity levels per architecture. For TopK we used $k \in {3,5,10}$; for L1-regularized, we used penalties from $(5,10,20)$. Each SAE had 100 latents and was trained for 500 epochs. We observed that low regularization tended to produce overly dense features, while increased regularization led to large amounts of dead features. These challenges were particularly difficult to balance with the limited settings, where the smaller number of features (100) and smaller dataset made training less stable and led to training that was more sensitive to hyperparameter choice. To address this, we manually identified a reasonable lower and upper bound for the hyperparameters that didn't create an obvious collapse of training.

We saved 10 checkpoints per run and evaluated them on a held-out validation set constructed identically to the training set, with four sentences per concept. To select the best checkpoint, we combined two metrics: (1) the average number of nonzero activations (to capture sparsity) and (2) downstream loss, measured by the KL divergence between the original model logits and the logits after injecting SAE reconstructions \cite{gao2024scalingevaluatingsparseautoencoders}. We normalized both metrics and combined them to form the selection objective. For Gemma, giving weight to sparsity led to a large amount of dead features, so only downstream loss was used for checkpoint selection. Overall, we trained 252 SAEs and evaluated 2520 checkpoints to identify the SAEs used.

\section{Comparison of Interpretability Results Across Four K Values}
\label{apx:interp_varying_k}

Here we describe the experimental settings of the concept detection benchmarking (\S\ref{sec:interp}) and provide comparisons for varying values of $k$.

\subsection{SNMF Dataset}
To construct the \texttt{SNMF} training data we randomly sampled 20 concepts from Neuronpedia \cite{neuronpedia}. For each concept we prompted GPT-4o-mini \cite{openai2024gpt4ocard} to generate 10 unique sentences exemplifying the concept. The resulting dataset contains 200 sentences exemplifying 20 labeled concepts and additional non-labeled concepts. For the prompt used to generate the dataset refer to \S\ref{apx:prompt_dataset_generation}.

\subsection{SNMF Training}
We applied \texttt{SNMF} as described in \S\ref{apx:nmf_details} with varying values of $k =100,200,300,400$. For Llama-3.1-8B and Gemma-2-2B, we kept the top 1 percent of indices in each MLP feature during training as described in \S\ref{apx:nmf_details}, whereas for GPT-2 we keep the top 5 percent due to its smaller dimensions. 

\subsection{SAE Concepts}
For the SAEs, we randomly sampled 100 features per layer from those identified by Neuronpedia \cite{neuronpedia}, using the SAEs from Llamascope, Gemmascope, and OpenAI’s GPT-2 Small \cite{he2024llamascope, lieberum2024gemmascopeopensparse, gao2024scalingevaluatingsparseautoencoders}, all trained on MLP-out.
\subsection{Evaluation}
The evaluation procedure is identical across all methods and layers.
For each concept and layer, we follow the same methodology: generate 5 activating sentences that exemplify the concept and 5 neutral sentences that do not. Each sentence is passed through the model, and we extract the activations at the target layer. For each activation, we compute the cosine similarity with the feature and record the maximum similarity across all tokens in the sentence.
We then average these maximum scores across the activating sentences to obtain $\bar{a}{\text{activating}}$, and across the neutral sentences to obtain $\bar{a}{\text{neutral}}$.
Finally, we compute the Concept Detection score ($S_{CD}$) as the log-ratio between the two averages:
\begin{equation}
S_{CD} := \log \frac{\bar{a}{\text{activating}}}{\bar{a}\text{neutral}}
\end{equation}

\subsection{Results}

\begin{figure}[H]
  \centering
  \includegraphics[width=0.9\columnwidth]{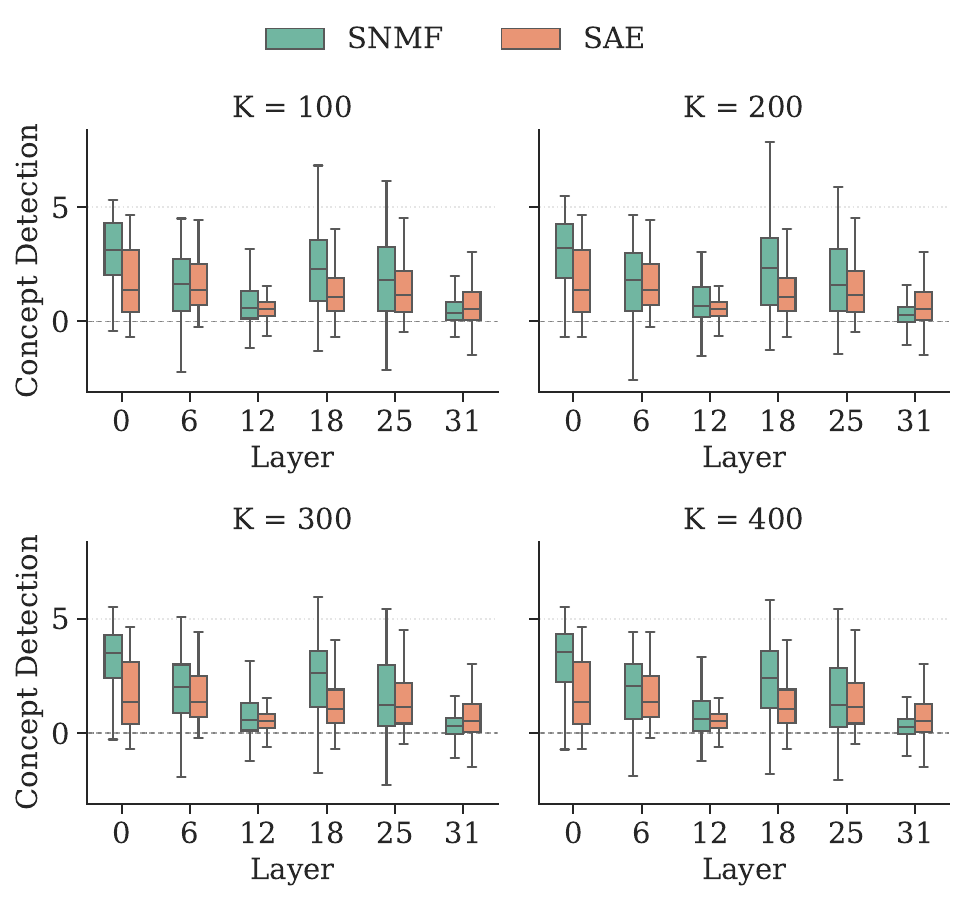}
  \caption{Interpretability scores for Llama-3.1-8B across \(K = 100, 200, 300, 400\).}
  \label{fig:app:interp_k_llama}
\end{figure}

\begin{figure}[H]
  \centering
  \includegraphics[width=0.9\columnwidth]{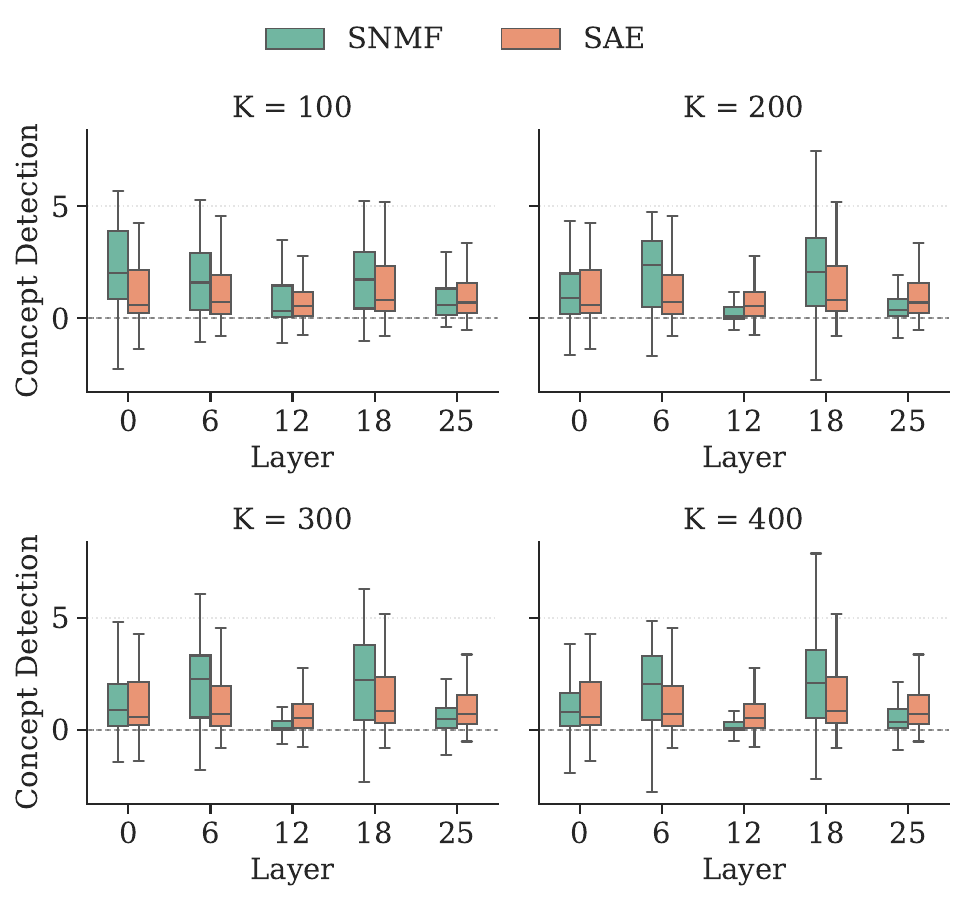}
  \caption{Interpretability scores for Gemma-2-2B across \(K = 100, 200, 300, 400\).}
  \label{fig:app:interp_k_gemma}
\end{figure}

\begin{figure}[H]
  \centering
  \includegraphics[width=0.9\columnwidth]{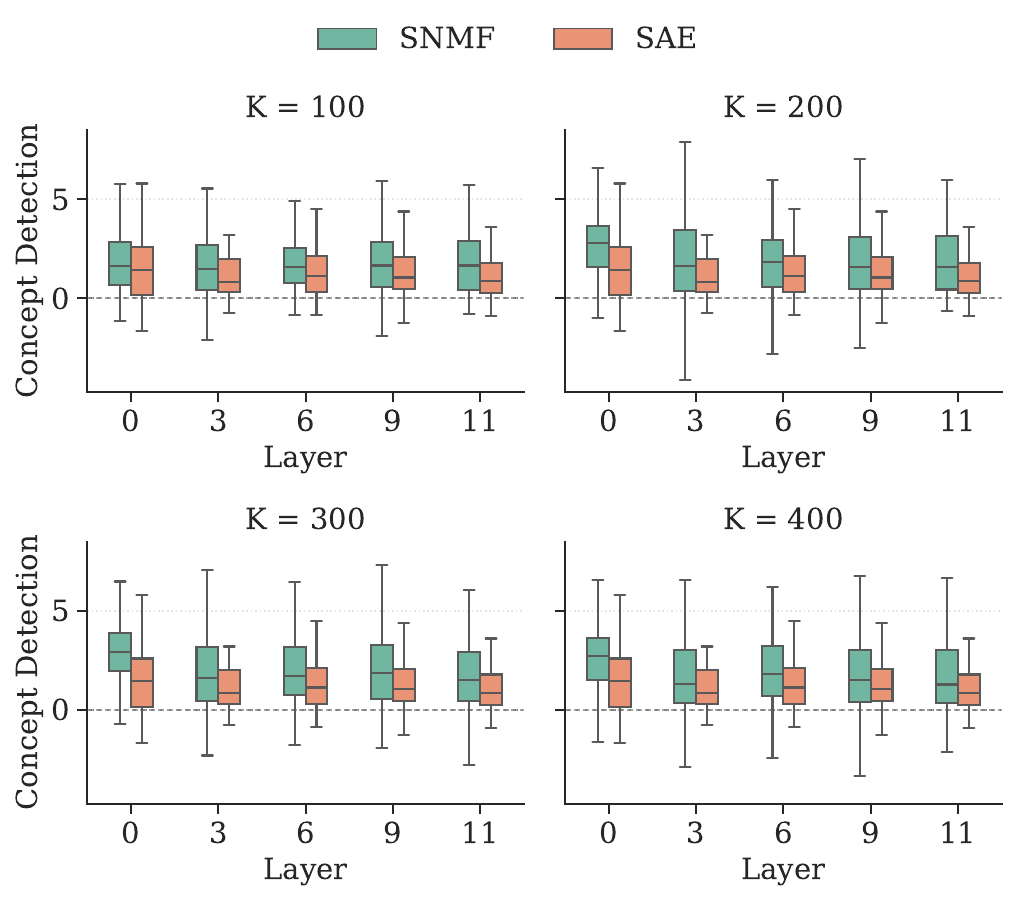}
  \caption{Interpretability scores for GPT-2 Small across \(K = 100, 200, 300, 400\).}
  \label{fig:app:interp_k_gpt2}
\end{figure}

Fig.~\ref{fig:app:interp_k_llama}, \ref{fig:app:interp_k_gemma} and \ref{fig:app:interp_k_gpt2}, present results for varying values of $k$ in \texttt{SNMF}, while for SAE the number of concepts is fixed at 100. Across all models, we observe that increasing $k$ generally improves cosine similarity in most layers, suggesting that larger \texttt{SNMF} decompositions yield features that more cleanly activate for specific concepts.
However, there are exceptions. For instance, in Gemma-2-2B at layer 12, performance degrades as $k$ increases. This may indicate that setting $k$ higher than the number of meaningful concepts present at a given layer can hurt performance.

\section{Additional Details on Causal Benchmark}
\subsection{Comparison to Released Causal Benchmarks}
 We designed the steering evaluation to be aligned with AxBench \cite{wu2025axbenchsteeringllmssimple}: measuring steering success by balancing target concept strength with fluency (instruction following is omitted as we do not evaluate instruction-tuned models). However, AxBench assumes a shared predefined concept set, while unsupervised methods such as SNMF and SAEs generally discover different features rather than the same fixed concepts. We therefore stay as close as possible to their described steering setup, prompts, and scoring, while adapting the concept definition and generation process to maintain a fair symmetric comparison across methods.

\subsection{Fluency Scores}
\begin{figure}[H]
  \centering
  \includegraphics[width=0.95\columnwidth]{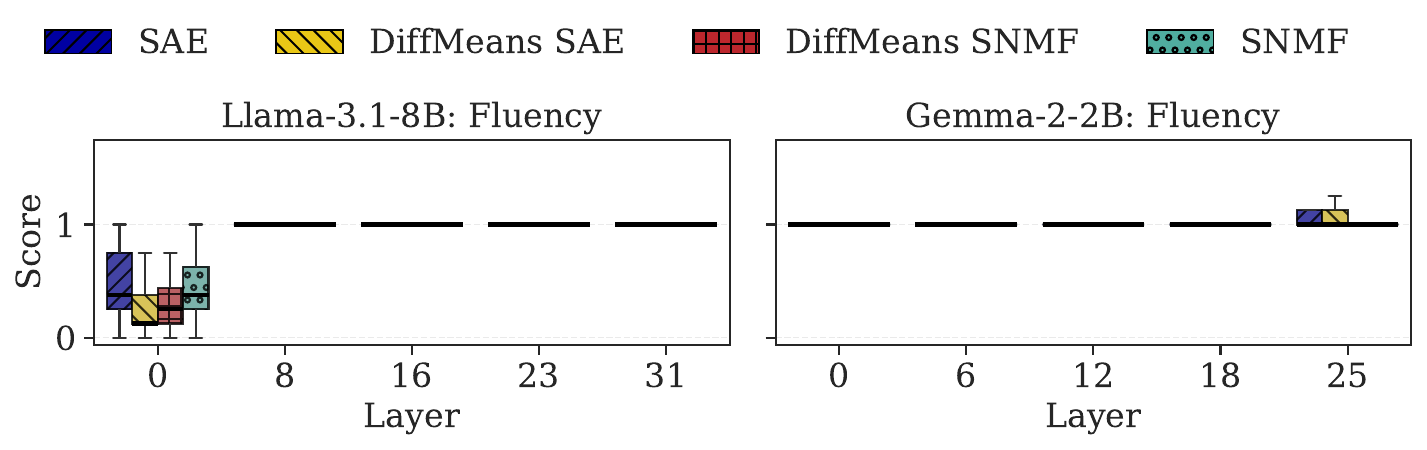}
  \caption{Fluency scores for Llama-3.1-8B and Gemma-2-2B across all methods: SAE, \texttt{DiffMeans} trained on SAE concepts, \texttt{DiffMeans} trained on SNMF concepts and SNMF.}
  \label{fig:app:fluency_scores}
\end{figure}

Fig.~\ref{fig:app:fluency_scores} presents the Fluency score results from \S\ref{sec:steering}. Notably, we see a low score for Layer 0 showing the increased difficulty in maintaining model coherence at lower layer interventions.

\label{apx:details_causality}

\section{Experimental Details of Neuron Compositionality}
\label{apx:neuron_comp}

\subsection{Recursive SNMF Algorithm}

We follow the method outlined in \citet{hier_nmf}. Initially, we decompose the activations as described in \S\ref{apx:nmf_details} to obtain matrices $Y_0$ and $Z_0$. Then we recursively decompose the matrix $Z_i$ to produce matrices $Y_{i+1}$ and $Z_{i+1}$. To improve the integration of the different MLP feature hierarchies, following \citet{hier_nmf}, we jointly fine-tune all layers using gradient descent to minimize the reconstruction loss between the original activations $A$ and their multi-layer approximation:
\[
\mathcal{L} = \frac{1}{2} \left\| A - Z_L Y_{L} \cdots Y_1 Y_0 \right\|_F^2
\]
The gradients are propagated backward through the factorization layers, and each $Z_i$ and $Y_i$ is updated using gradient descent. This yields a set of multi-level MLP features, where higher-layer features encode increasingly abstract combinations of co-activated neurons.

\subsection{Dataset Generation: Time Units}
We prompted GPT-o4-mini-high to generate a dataset focused on the concept of units of time: years, months, weekdays, minutes etc. We chose this concept because prior work \cite{engels2025languagemodelfeaturesonedimensionally} has shown that GPT-2 represents weekdays and months in a multi-dimensional manner, where each day of the week is positioned near the ones that follow it, forming a circular structure in a lower-dimensional subspace. We use the following prompt:

\tcbset{
    left*=10pt, right*=10pt,
    top=0pt, bottom=0pt,
    colback=white!10!white,
    colframe=black!75!black,
    fonttitle=\bfseries\large,
    subtitle style={boxrule=0pt,colback=gray!50!white},
}
\begin{tcolorbox}[title=Dataset Generation Prompt]
\quad \\
\small
    Generate a dataset of sentences that are all related to days of the week, months, years, dates etc. Make sure to include plenty of both broad and specific terms including all names of the days and months and for each term over 10 instances. Generate 200 sentences. Format the dataset with two columns: Prompt, Label. Format as a json in the format {<label>: [<sentences>]}.
\end{tcolorbox}

\subsection{Dataset Generation: General}
We constructed a more general dataset by prompting GPT-4o for 50 random nouns as base concepts. For each noun, we queried ConceptNet to retrieve related concepts that represent subcomponents or subcategories using relations like HasA, PartOf, or TypeOf. For each of the related concepts, we prompted Gemini-2.0-flash \cite{google2025gemini} to generate five distinct sentences exemplifying the relationship in natural language. This process resulted in a total of 873 sentences that reflect a diverse range of semantically related sentence groups.

\subsection{Tree Construction}

To identify hierarchical MLP features, we apply recursive \texttt{SNMF} with $K = [400, 200, 100, 50]$. Each matrix $Y_i$ maps features from level $i{+}1$ to those in $Z_i$, allowing us to trace how higher-level features activate lower-level ones. To construct the tree, we connect each feature at level $i{+}1$ to its top-activated features in $Z_i$ according to $Y_i$, using a fixed threshold to prune weak edges. This forms a directed tree structure that reflects the compositional organization of features across levels. We label the nodes corresponding to the MLP features using the top activating contexts associated with the MLP features. We obtain the mapping for an MLP feature at layer i in the hierarchy to the contexts by multiplying the coefficient matrices $0$ to $i$: $Y^{(0 \rightarrow i)} = Y_0 Y_1 \cdots Y_i$

\subsection{Additional Hierarchical Decompositions}

\begin{figure}[H]
  \includegraphics[width=0.7\columnwidth]{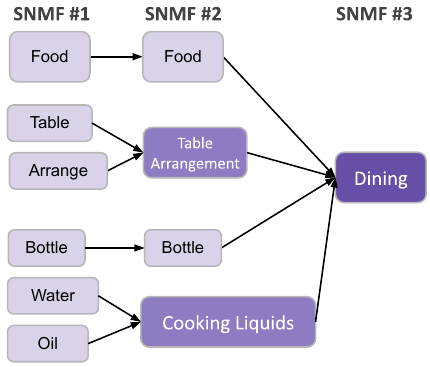}
  \caption{Merging of Dining related MLP features.
  \label{fig:app:tree1}
  }
\end{figure}

\begin{figure}[H]
  \includegraphics[width=0.7\columnwidth]{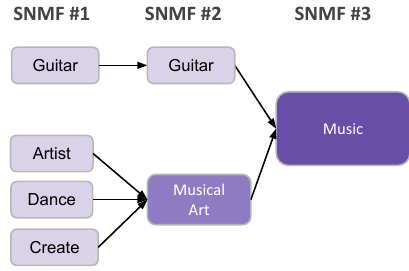}
  \caption{Merging of Music related MLP features.
  \label{fig:app:tree2}
  }
\end{figure}

\begin{figure}[H]
  \includegraphics[width=0.7\columnwidth]{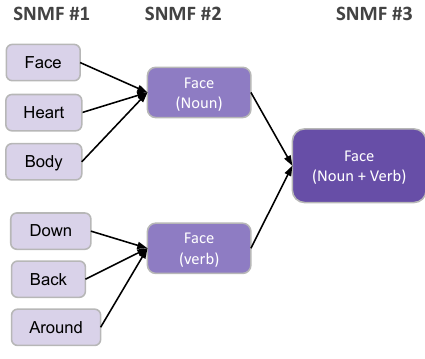}
  \caption{Hierarchy showing that semantically different, but syntactically related features exhibit merging as well.
  }
  \label{fig:app:tree3}
\end{figure}

\begin{figure}[H]
  \includegraphics[width=0.55\columnwidth]{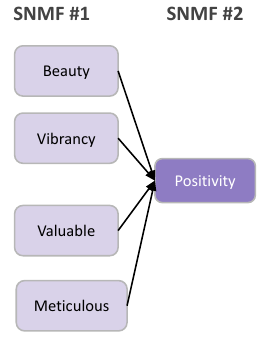}
  \caption{Hierarchy showing that semantically related features share neurons.
  \label{fig:app:tree4}
  }
\end{figure}

\begin{figure}[H]
  \includegraphics[width=0.7\columnwidth]{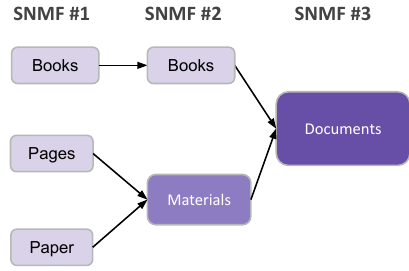}
  \caption{Hierarchy showing that compositional concepts exhibit neuron sharing.
  \label{fig:app:tree5}
  }
\end{figure}

In Fig.~\ref{fig:app:tree1}, \ref{fig:app:tree2}, \ref{fig:app:tree3}, \ref{fig:app:tree4} and \ref{fig:app:tree5} we provide additional examples of hierarchical MLP features found by applying the described methodology on the general dataset on MLP activations extracted from Layer 0. We manually select the examples to provide a variety of compositions. We note that these are a select few from a larger pool of identified hierarchical MLP features.


\begin{table*}[h!]
\centering

\begin{tabular}{lcccc}
\toprule
{Metric} & \multicolumn{2}{c}{Llama-3.1\,8B ($n=93$)} & \multicolumn{2}{c}{Gemma-2\,2B ($n=67$)}\\
\cmidrule(lr){2-3}\cmidrule(lr){4-5}
 & Ours & Random & Ours & Random \\
\midrule
Coverage & $0.46 \pm 0.20$ & $0.05 \pm 0.09$ & $0.37 \pm 0.18$ & $0.00 \pm 0.00$ \\
Diversity & $0.99 \pm 0.04$ & $1.00 \pm 0.01$ & $0.99 \pm 0.03$ & $1.00 \pm 0.00$ \\
Mean cov/child & $0.47 \pm 0.20$ & $0.05 \pm 0.09$ & $0.38 \pm 0.18$ & $0.00 \pm 0.00$ \\
\bottomrule
\end{tabular}
\caption{%
Parent--child feature analysis for two models (mean $\pm$ std).
\textbf{Coverage}: fraction of tokens promoted by any child feature that are also promoted by its parent.
\textbf{Diversity}: $1-$overlap among children’s top tokens (higher means children specialize in different sub-concepts).
\textbf{Mean cov/child}: average coverage when each child is considered individually.
\(n\) = number of parent features evaluated.
\textbf{Random} = baseline obtained by pairing each parent with the same number of randomly chosen lower-level features.
}
\label{tab:parent_child}
\end{table*}

\section{Hierarchal Features in MLP Layers}
\label{apx:hierarchy}

To assess whether the identified hierarchy is a property of the MLP, we conducted a broader experiment to evaluate whether a parent feature’s causal effect covers the causal effects of its children in their hierarchy. Similar to the phenomenon observed in the weekday example (Table~\ref{tab:causal_effect_exclusive_neurons}).

First, we built a test set with hierarchical concepts. Using GPT-4o, we sampled 50 random “base” nouns, such as “car". For each noun, we queried ConceptNet for HasA, PartOf, SubstanceOf and MemberOf relations, yielding finer-grained sub-concepts (e.g., “a tire” is a part of a car). Then we utilized Gemini-2.0-flash to generate five sentences exemplifying each sub-concept. Overall, we generated a dataset of 873 sentences capturing various hierarchies in language. Next, we obtained activations for these examples and passed them through a two-level recursive SNMF pipeline (ranks K=500, 200) and automatically extracted parent-child links. We consider a lower level feature a child when it contributes at least $2.5\%$ of its parent’s column mass which is distributed over 500 features in this setting.

For every parent with at least two children, we measured \texttt{coverage} as follows: For a fixed set of
20 initial prompts, we computed the model’s logits with and without intervention, intervening with
each feature (parent and children) in isolation. For each intervention, we identified the top 50
tokens whose logits increased the most with respect to the non-intervened logits. We then
computed \texttt{coverage} as the fraction of the children’s promoted tokens (the union of their top-10
sets) that also appeared in the parent’s top-50 set. Additionally, we measure the \texttt{coverage} of
each parent with a single child and average across the children in order to validate that the
\texttt{coverage} is not coming from a single child feature, we term this metric \texttt{mean coverage per child}.
Together, these metrics quantify how well the parent captures the combined semantic
effect of its children.

We report the results for both Gemma-2 2B and Llama-3.1 8B in Table~\ref{tab:parent_child}. For both models, the average \texttt{coverage} was considerably higher than a control baseline, where we randomly selected unrelated features (rather than true parents) and computed their coverage using the same procedure. At the same time, the overlap among children (intersection of children's top 10 promoted logits divided by union of children's top 10 promoted logits) stayed low, confirming that children specialize in different sub-concepts. Lastly, the mean coverage per child was within 0.02 of the \texttt{coverage}, showing the parent is semantically related to each of the children. Since parents consistently cover the union of their children’s effects while the children remain diverse, the results provide strong evidence that the broad-to-specific hierarchies we observe are systematic properties of the MLP activation space, not artifacts of the SNMF decomposition.

\section{Concept Detection Prompts}

In this section we present the prompts used for the concept detection experiments.

\subsection{Generating Dataset From Concept Descriptions}
\label{apx:prompt_dataset_generation}

The following prompt is used sequentially to generate the dataset employed in identifying MLP features with \texttt{SNMF} in Section~\ref{sec:interp} and Section~\ref{sec:steering}. To generate the dataset, we provide the concept and prompt the model $m$ times, where $m$ is the desired number of sentences per concept.

\tcbset{
    left*=10pt, right*=10pt,
    top=0pt, bottom=0pt,
    colback=white!10!white,
    colframe=black!75!black,
    fonttitle=\bfseries\large,
    subtitle style={boxrule=0pt,colback=gray!50!white},
}
\begin{tcolorbox}[title=Dataset Generation Prompt]
\quad \\
\small
    Generate a sentence that exemplifies and strongly exhibits the concept of \texttt{\{concept\}}.\\
Make sure the sentence is different from the following list: \texttt{\{current\_sentences\}}.\\
Output the sentence \textbf{ONLY} without additional text.
\end{tcolorbox}

\subsection{Generating Activating and Neutral Sentences}
\label{apx:prompt_act_neutral}

The following prompts are used to generate sentences that exemplify a concept (activating sentences) and those that do not (neutral sentences). To construct the list, we sequentially prompt the model to generate a new sentence while conditioning on the current list to encourage diversity.

\tcbset{
    left*=10pt, right*=10pt,
    top=0pt, bottom=0pt,
    colback=white!10!white,
    colframe=black!75!black,
    fonttitle=\bfseries\large,
    subtitle style={boxrule=0pt,colback=gray!50!white},
}

\begin{tcolorbox}[title=Activating Prompt]
You are given a description of an LLM concept.\\
Given the description, generate a sentence that contains tokens that would activate the feature. Make sure your generated sentence exemplifies all the key terms and structures specified in the description.\\
Ensure that the sentence is a full, grammatically correct English sentence.

\textbf{Category description:}\\
\texttt{\{description\}}

Make sure you generate a sentence that is distinct from: \texttt{\{previous\_sentences\}}\\

Output only the sentence without any additional text.
\end{tcolorbox}

\begin{tcolorbox}[title=Neutral Prompt]
You are given a description of an LLM concept.\\
Your objective is to generate a neutral sentence that should not activate the feature.\\
This means that you must not include in the sentence any tokens that relate to the feature.\\
Have the generated sentence be on a completely unrelated topic.

\textbf{Concept description:}\\
\texttt{\{description\}}

Make sure you generate a sentence that is distinct from: \texttt{\{previous\_sentences\}}\\

Output only the sentence without any additional text.\\
Make sure to exclude any tokens that may activate the concept.
\end{tcolorbox}

\subsection{Input Description Prompt}

The following prompt is used to generate a natural language description for each concept, based on the top activated tokens in the matrix $Y$, which represents token-level activations for each MLP feature. Specifically, for a given MLP feature, we extract the tokens with the top $m$ highest activations and use them as input to the prompt. The model is asked to interpret the shared semantic content of these tokens. We then parse the \textit{Results} section of the model's response to retrieve the final concept description.

\begin{figure*}
\input{prompt_input_desc_gen_prompt}
\end{figure*}

\end{document}